\documentclass[letterpaper]{article} 
\usepackage{aaai25}  
\usepackage{times}  
\usepackage{helvet}  
\usepackage{courier}  
\usepackage[hyphens]{url}  
\usepackage{graphicx} 
\usepackage{natbib}  
\usepackage{caption} 
\usepackage{booktabs} 
\usepackage{adjustbox}
\usepackage{amssymb} 
\usepackage[table,dvipsnames]{xcolor}
\usepackage{fontawesome} 
\usepackage{amsmath} 
\usepackage{algpseudocode}
\usepackage{subcaption}
\usepackage{algorithm}
\usepackage{newfloat}
\usepackage{listings}
\usepackage{enumitem}
\usepackage[dvipsnames]{xcolor}
\DeclareCaptionStyle{ruled}{labelfont=normalfont,labelsep=colon,strut=off} 
\floatstyle{ruled}
\newfloat{listing}{tb}{lst}{}
\floatname{listing}{Listing}

\title{Measuring Cross-Modal Interactions in Multimodal Models}

\author {
    Laura Wenderoth\textsuperscript{\rm 1},
    Konstantin Hemker\textsuperscript{\rm 1},
    Nikola Simidjievski\textsuperscript{\rm 2,1},
    Mateja Jamnik\textsuperscript{\rm 1}
}
\affiliations {
    \textsuperscript{\rm 1}Department of Computer Science and Technology, University of Cambridge, Cambridge, United Kingdom\\
    \textsuperscript{\rm 2}PBCI, Department of Oncology, University of Cambridge, Cambridge, United Kingdom\\
    \{lw457, kh701, ns779, mj201\}@cam.ac.uk
}

\begin{document}

\maketitle

\begin{abstract}
Integrating AI in healthcare can greatly improve patient care and system efficiency. However, the lack of explainability in AI systems (XAI) hinders their clinical adoption, especially in multimodal settings that use increasingly complex model architectures. Most existing XAI methods focus on unimodal models, which fail to capture cross-modal interactions crucial for understanding the combined impact of multiple data sources. Existing methods for quantifying cross-modal interactions are limited to two modalities, rely on labelled data, and depend on model performance. This is problematic in healthcare, where XAI must handle multiple data sources and provide individualised explanations. This paper introduces InterSHAP, a cross-modal interaction score that addresses the limitations of existing approaches. InterSHAP uses the Shapley interaction index to precisely separate and quantify the contributions of the individual modalities and their interactions without approximations. By integrating an open-source implementation with the SHAP package, we enhance reproducibility and ease of use. We show that InterSHAP accurately measures the presence of cross-modal interactions, can handle multiple modalities, and provides detailed explanations at a local level for individual samples. Furthermore, we apply InterSHAP to multimodal medical datasets and demonstrate its applicability for individualised explanations. \looseness -1

\end{abstract}

\begin{links}
\link{Code}{github.com/LauraWenderoth/InterSHAP}
\link{Extended version}{arxiv.org/abs/2412.15828}
\end{links}

\section{Introduction}
\begin{figure}[t!]
    \centering
    \includegraphics[width=0.48\textwidth]{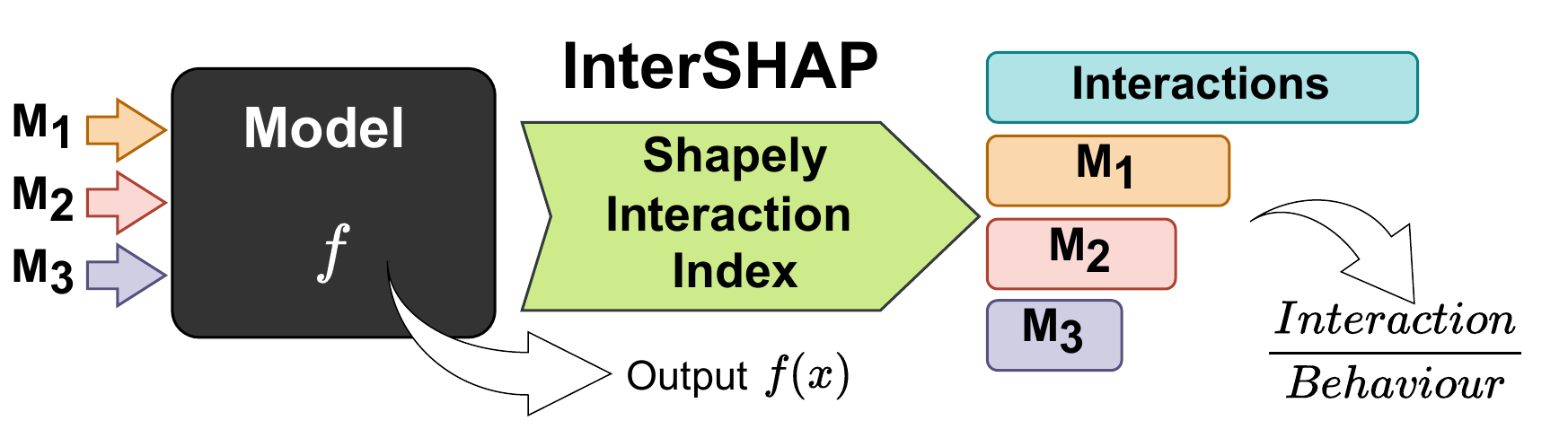}
    \caption[Schematic overview of InterSHAP]{Overview of InterSHAP. The model (black box), takes three different modalities as input and produces an output \( f(x) \). Through perturbations of the input modalities and observing the resulting changes in outputs, the Shapley interaction index \cite{grabisch_axiomatic_1999} is used to dissect the model's behaviour into modality contributions and cross-modal interactions. \textit{InterSHAP is defined as the ratio of interactions to model behaviour}.}
    \label{fig:overview_fig}
    \vspace*{-2mm}
\end{figure}

In medical decision-making, there is a growing trend toward utilising multimodal machine learning approaches that integrate multiple data sources to improve predictive and diagnostic tasks. These approaches acknowledge that medical data analysis is inherently multimodal, leveraging methods that are able to fuse heterogeneous data that includes clinical, genomic, and imaging modalities. Recent advancements in large multimodal models in the medical field, such as Google's Med-PaLM \cite{tu_towards_2023} and Microsoft's BiomedCLIP \cite{zhang_biomedclip_2023}, further highlight the utility of multimodal approaches in healthcare. \looseness -1

However, an inherent limitation of many of the current multimodal approaches pertains to their explainability.  Thus, healthcare applications of machine learning are often left to rely on high-performance but opaque models that obscure the reasoning behind their predictions. Such models can accurately predict diagnoses, but these predictions may not reflect true causal relationships in the data \cite{han_sounds_2022}.
The lack of explainability and transparency is widely recognized as a key barrier to the clinical adoption of ML models, as noted by policymakers such as the OECD \cite{anderson_collective_2024} and emphasized in academic studies \cite{yang_unbox_2022}.

In response to the urgent need for XAI, numerous methods such as SHAP \cite{lundberg_unified_2017} and LIME \cite{ribeiro_why_2016} have emerged to explain the results of machine learning systems. While these have shown promise for unimodal models, there are few explanation methods that work effectively across data structures, meaning that cross-modal interactions can often not be quantified with unimodal methods. Multiple modalities allow a machine learning model to utilise and combine information from different sources, enabling cross-modal interactions that often lead to improved performance. However, simply identifying these interactions is insufficient for a meaningful interpretation of the model predictions. In this context, it is crucial to comprehend both the individual impact of each modality and their combined influence. Several methods, such as {PID} \cite{liang_quantifying_2023}, {EMAP} \cite{hessel_does_2020}, and  {SHAPE} \cite{hu_shape_2022}, have been developed to assess whether models learn cross-modal interactions. However, these methods are often (1)~limited to only two modalities; (2)~applied to the entire dataset, rather than individual samples; and (3)~either require labelled data or are not performance-agnostic, which can lead to an incomplete understanding of cross-modal interactions. 

In this paper, we introduce \textbf{InterSHAP}, a comprehensive, interpretable metric that quantifies the influence of cross-modal interactions on model behaviour. InterSHAP supports any number of modalities, provides local explanations, is agnostic to model performance, and works with unlabelled data. To our knowledge, our proposed variant of the Shapley interaction index is the first approach that is able to effectively separate cross-modal interactions from individual modality contributions. Figure~\ref{fig:overview_fig} provides a schematic overview of InterSHAP. To validate the effectiveness of InterSHAP, we conduct extensive empirical analyses using synthetic datasets with either no synergy (no cross-modal interactions are needed to solve the task) or exclusively synergy (complete information can only be obtained as a result of cross-modal interaction across all modalities). The results demonstrate that InterSHAP can accurately detect a lack of cross-modal interactions (\(0\%\)) on datasets with no synergy, and \(99.7\%\) on those with exclusive synergy, aligning precisely with the expected behaviour. Furthermore, we demonstrated that \mbox{InterSHAP} is extendable to more than two modalities and achieves favourable results at a local level with individual data points. Additionally, we test InterSHAP on two medical datasets to show its utility in a real-world setting. InterSHAP is not confined to medical domains and can be applied to models trained on any modality.

Our main contributions can be summarised as follows:
\begin{itemize}[itemsep=-2pt,leftmargin=*]
    \item \textbf{Novel cross-modal interaction score  InterSHAP.} We introduce InterSHAP, the first model-agnostic cross-modal interaction score that operates with any number of modalities, offering both local and global explanations, applies to unlabelled datasets and does not rely on the model's performance.
    We conducted extensive experiments on synthetic datasets and models to validate its functionality. \looseness -1
    \item \textbf{Application to multimodal healthcare datasets.} We demonstrate the benefits of InterSHAP on a cell type classification task based on multimodal protein and RNA data as well as a mortality prediction tasks.
    \item \textbf{Open-source implementation with integration into the SHAP Package.} We have developed an open-source implementation of InterSHAP that seamlessly integrates with the well-known SHAP visualisation package \cite{lundberg_unified_2017}.

\end{itemize}

\section{Related Work}
\begin{table}[t]

\centering
\begin{adjustbox}{width=1\linewidth}
\begin{tabular}{lcccccc} 
\toprule
Score &   \shortstack{Modalities\\$> 2$}  & Local & Unsupervised& \shortstack{Performance \\ Agnostic}  & 
\\
\midrule
PID & \textcolor{Mahogany}{\faTimes} & \textcolor{Mahogany}{\faTimes} & \textcolor{Green}{\checkmark} & \textcolor{Green}{\checkmark} 
\\
EMAP & \textcolor{Mahogany}{\faTimes} & \textcolor{Mahogany}{\faTimes} & \textcolor{Mahogany}{\faTimes} & \textcolor{Mahogany}{\faTimes} 
\\
SHAPE & \textcolor{Green}{\checkmark} & \textcolor{Mahogany}{\faTimes} & \textcolor{Mahogany}{\faTimes} & \textcolor{Mahogany}{\faTimes} 
\\[1mm]
\textbf{InterSHAP}& \textcolor{Green}{\checkmark} & \textcolor{Green}{\checkmark} & \textcolor{Green}{\checkmark} & \textcolor{Green}{\checkmark} 
\\
\bottomrule
\end{tabular}
\end{adjustbox}

\caption{InterSHAP overcomes the limitations of other cross-modal interaction scores: it is unsupervised, performance agnostic, applicable to more than two modalities, and allows for dataset- (global) and sample-level (local) explainability. }
\label{tab:limitations of cross modal interaction scores}
\vspace*{-3mm}

\end{table} 
Existing approaches like MM-SHAP \cite{parcalabescu_mm-shap_2023} and Perceptual Score \cite{gat_perceptual_2021} focus on understanding the contributions of individual modalities rather than explicitly examining their interactions. They explain modality contributions in isolation without considering or quantifying cross-modal interactions within multimodal networks.
In contrast, EMAP \cite{hessel_does_2020} and SHAPE \cite{hu_shape_2022} detect cross-modal interactions by analysing how perturbations to the input impact the model’s output. They systematically vary different modalities or combinations of modalities to assess their importance within the model. 
Another metric for detecting interactions, Partial Information Decomposition (PID) \cite{liang_quantifying_2023}, quantifies synergy (interactions between modalities in a dataset with a combined effect surpassing individual contributions, indicating a non-additive relationship) in datasets, which can be interpreted as cross-modal interactions. \looseness -1

However, these existing approaches exhibit various limitations, summarised in Table~\ref{tab:limitations of cross modal interaction scores}. First, most of them restrict the analysis to just two modalities, which prevents them from being applied in domains that typically have more modalities, like in healthcare. 
Second, the results cannot be analysed at the level of individual samples, which is particularly critical for medical tasks where explainability for each patient is crucial. 
Third, some approaches necessitate the availability of labels, rendering them unsuitable for unlabelled datasets employed in self- or unsupervised learning scenarios. 
Fourth, certain methods assess interactions solely based on performance gain, rendering them highly performance-dependent. However, when assessing a model, it is crucial to understand how each interaction contributed to the present prediction without neglecting interactions only because the prediction was incorrect. \looseness -1

\section{Methodology}
We introduce InterSHAP to quantify cross-modal interaction learned by multimodal models while overcoming the limitations of SOTA explainability approaches. 
Following the definition of \citeauthor{liang_foundations_2022}~\shortcite{liang_foundations_2022}, we define cross-modal interaction  as a change in the model response that cannot be attributed to the information gained from a single modality alone, but arises only when both modalities are present.
To ensure the accuracy of InterSHAP, we use the Shapley Interaction Index (SII), which has been shown to effectively capture model behaviour \cite{lundberg_local_2020} and decompose unimodal feature interactions into its constituent parts \cite{ittner_feature_2021}.

\subsection{Preliminaries}

\paragraph{Model Fusion.}
To train multimodal models is model fusion, which comes in three types: (i)~early fusion, which merges data features at model input, (ii)~intermediate fusion, which combines data within the model, and (iii)~late fusion, which combines the outputs of models trained on different individual data modalities \cite{stahlschmidt_multimodal_2022,azam_review_2022}. 

\paragraph{Shapley Value.} 
The Shapley value \(\phi\), from cooperative game theory, measures each feature's contribution to overall performance via marginal contributions to feature coalitions~\cite{roth_introduction_1988}. In machine learning, a model \(f_M\) is viewed as a coalition game with \(M\) features or modalities as players. To avoid inefficient retraining, masking strategies are applied to features \(j \notin S\), where \(S \subseteq M\). 

The Shapley value $\phi$ of a feature \(i\) is calculated as the weighted average of their marginal contributions to all possible coalitions:
\begin{equation}
\label{eq:shapley value}
\begin{aligned}
\phi_i (M,f) &= \sum_{S\subseteq M \backslash\{i\}} \frac{|S| !(|M|-|S|-1) !}{|M| !} \Delta \\
\Delta &= \left[f_{S \cup\{i\}}\left(S \cup\{i\}\right) - f_S\left(S\right)\right].
\end{aligned}
\end{equation}

\paragraph{Shapley Interaction Index.} 
The Shapley Interaction Index (SII) \cite{grabisch_axiomatic_1999}, initially devised to quantify interaction effects between players in cooperative game theory, can be employed to discern interactions between features or modalities. We define its directly adapted version  $\phi_{ij}(M,f)$ for $M$ modalities and a model \(f\), where \(i,j\in M\)~\cite{lundberg_local_2020} and  \(i \neq j \):
\begin{equation}
\label{eq:shapleyinteractionindex}
\phi_{ij}(M,f)=\!\!\sum_{S \subseteq M \backslash \{i, j\}} \!\!\frac{|S| !(M-|S|-2) !}{2(M-1) !} \nabla{i j}(S,f), 
\end{equation}
\begin{equation}
\begin{aligned}
\nabla_{ij}(S,f) = & \left[f_{S \cup\{ij\}}\left({S \cup\{ij\}}\right)-f_{S \cup\{i\}}\left({S \cup\{i\}}\right) \right. \\
& \left. - f_{S \cup\{j\}}\left({S \cup\{j\}}\right) + f_S(S)\right].
\end{aligned}
\vspace{0.7em}
\end{equation}

The interaction index for a modality with itself, denoted as \(\phi_{ii}\), is defined as the difference between the Shapley value \(\phi_i\) and the sum of all interaction values:
\begin{equation}
    \phi_{ii}(M,f) = \phi_{i}(M,f) - \sum_{j \in M} \phi_{i j}(M,f) \quad \forall i \neq j.
    \label{eq:interaction_index_ii}
\end{equation}

\subsection{Global InterSHAP}  
We aim to quantify the impact of cross-modal interactions on model behaviour using InterSHAP. Model behaviour is a cumulative process, beginning with the model’s base output and progressively adding each modality and their interactions to shape the final prediction.
To isolate interactions from the model behaviour, we apply the SII. Let \( f \in \mathbb{R}^c \) be a trained model, where \( c \) is the number of per-class probabilities, evaluated on a dataset with \( N \) samples and \( M \) modalities, represented as \( \{(m_1^i, \dots, m_M^i)\}_{i=1}^N \) and corresponding labels \( \{y^i\}_{i=1}^N \). The SII value \( \phi_{ij} \) measures the interaction between modalities \( i \) and \( j \) for sample \( a \). \(\Phi_{ij}\) is defined as the absolute value of the mean over \(\phi_{ij}\) of all samples in the dataset: \looseness -1
\begin{equation}
\label{eq:interaction value}
\Phi_{ij} = \left| \frac{1}{N} \sum_{a=1}^N \phi_{ij}(m_1^a, \dots, m_M^a, f) \right|,
\end{equation}  
resulting in the matrix  $\Phi$ containing only positive values:
\begin{equation}
    \Phi = \begin{bmatrix}
        \Phi_{11} & \textcolor{Green}{\Phi_{12}} & \dots & \textcolor{Green}{\Phi_{1M}} \\
        \textcolor{Green}{\Phi_{21}} & \Phi_{22} & \dots & \textcolor{Green}{\Phi_{2M}} \\
        \vdots & \vdots & \ddots & \vdots \\
        \textcolor{Green}{\Phi_{M1}} & \textcolor{Green}{\Phi_{M2}} & \dots & \Phi_{MM}
    \end{bmatrix},
\end{equation}  
where interactions appear in green, while modality-specific contributions are shown in black. Note that, our choice of obtaining $\Phi_{ij}$ allows for a global interpretability view, accounting for opposing effects that may cancel out across dataset. This prevents score inflation and aligns with our main motivation -- a general and robust measure of interaction strength. In turn, these are aggregated to obtain the total interactions contributions as well as the overall model behaviour: 
\begin{equation}
\label{eq:interactions}
    {Interactions} = \sum_{\substack{i, j = 1 \\ i \neq j}}^{M} \Phi_{ij}, \quad {Behavior} = \sum_{i, j = 1}^{M} \Phi_{ij}.
\end{equation}  
Finally, these are used to calculate InterSHAP, which is the fraction of model behaviour that can be explained through cross-modal interactions:
\begin{equation}
    InterSHAP = \frac{{Interactions}}{{Behavior}}.
    \label{eq:I dataset}
\end{equation}

\vspace*{-2mm}
\subsection{Modality Contributions} 
Based on the definition of cross-modal interactions in Equation~\ref{eq:interactions}, the contribution of all modalities $\mathcal{M}$ and the contribution of each modality $\mathcal{M}_i$ to the overall result is:
\begin{equation}
    \mathcal{M} = 1 - {InterSHAP}, \quad \mathcal{M}_i = \frac{\left|\Phi_{ii}\right|}{{Behaviour}}.
\end{equation}

\subsection{Local InterSHAP}
\label{Section:Methods local InterSHAP}
After determining the amount of interactions that a model $f$ has learned across the entire dataset, we assess the extent to which cross-modal interactions among the $M$ modalities contribute to the model's behaviour for each sample. Thus, we apply the definition of InterSHAP directly to a sample $a$:
\vspace*{-2mm}
\begin{equation}
    I_a = \frac{\sum_{\substack{i, j = 1 \\ i \neq j}}^{M} \varphi_{ij}^a}{ \sum_{i, j = 1}^{M} \varphi_{ij}^a },
\end{equation}

\vspace*{2mm}
\noindent
where $\varphi_{ij}^a\!=\!\left| \phi_{ij}(\left(m_1^a, \dots , m_M^a\right),f) \right|$, and $i, j \!\!\in\!\! \{1, \dots, M\}$.
To determine if InterSHAP is effective at the level of all individual data points, we aggregate $I_a$ across the entire dataset to obtain the local interaction score for the dataset as a whole:
\vspace{-2mm}
\begin{equation}
    InterSHAP_{local} = \frac{1}{N} \sum_{a=1}^N I_a.
\end{equation}

\begin{table*}[t!]

    \centering
        \begin{tabular}{lrrrrrrrrrr} 
            \toprule
            & \multicolumn{2}{c}{$\text{Uniqueness}$}  & \multicolumn{2}{c}{Synergy} & \multicolumn{1}{c}{Redundancy} & \multicolumn{1}{c}{Random} \\
            \cmidrule(lr){2-3} \cmidrule(lr){4-5} \cmidrule(lr){6-6} \cmidrule(lr){7-7}   
           &  XOR & FCNN & XOR & FCNN & FCNN & FCNN  \\
            \midrule
            ${\text{InterSHAP}}$  &0.0 & 0.2 $_{\pm  0.1}$ & 99.7 & 98.0 $_{\pm  0.5}$ & 38.6 $_{\pm  0.5}$ & 57.8 $_{\pm  1.1}$ \\
            ${\text{InterSHAP}_{local}}$  & 1.8 & 3.4 $_{\pm  5.2}$ &  96.9 & 85.8 $_{\pm  12.1}$ & 37.3 $_{\pm  25.0}$ & 40.0 $_{\pm  13.3}$ \\
            \midrule
            $\text{PID}$ &   0.01 & 0.01 $_{\pm  0.01}$ & 0.39 & 0.39 $_{\pm  0.01}$& 0.14 $_{\pm  0.02}$& 0.48 $_{\pm  0.01}$\\

EMAP$_{gap}$ & 0 & 0 $_{\pm  0}$ & 49.1 & 43.5 $_{\pm  1.0}$ & 1.8 $_{\pm  0.4}$ & 0.6 $_{\pm  0.4}$ \\
SHAPE &  1.6 & 16.5 $_{\pm  0.6}$ & 33.1 &  27.6 $_{\pm  1.5}$   &  -47.1  $_{\pm   0.5}$ &  15.1 $_{\pm   1.9}$  \\

            \bottomrule
        \end{tabular}

\caption[Local and global InterSHAP values in comparison]{Local and global InterSHAP values are presented in percentages for both the XOR function and the FCNN with early fusion on HD-XOR dataset with two modalities. EMAP$_{gap}$ (EMAP - F1 of the model) and SHAPE indicate F1 score improvement from cross-modal interactions, with negative values signalling performance decline. For PID, the synergy value is provided, though it is not expressed in any standard unit. For all metrics, higher scores indicate greater measured cross-modal interactions. Results for the XOR function align with expectations, confirming the effectiveness of the InterSHAP implementation.\looseness-1}
\label{tab:main_results}
\end{table*}

\subsection{Asymptotic Bound}
InterSHAP is characterised by a computational complexity of \(O(N^M)\), where \(N\) represents the number of samples and \(M\) signifies the number of modalities \cite{grabisch_axiomatic_1999,lundberg_local_2020}. The constraint of exponential growth emerges with the increasing number of modalities rather than the feature count. InterSHAP's dependency on the number of features within a modality follows a constant pattern, meaning its runtime remains unaffected by the feature count.

\section{Experimental Validation on Synthetic Data}

To verify InterSHAP's functionality, we control three factors influencing cross-modal interactions: dataset synergy (non-additive interactions between modalities), model learning capacity, and metric effectiveness.

\subsection{Setup}
\label{section:setup-experimental}
We utilise synthetic high-dimensional tabular datasets, generated using an XOR function (HD-XOR), with varying degrees of synergy to regulate the extent of cross-modal interactions. Each generated dataset contains 20,000 samples, with two to four modalities represented as 1D vectors consisting of around 100 features. The datasets are categorised on their degrees of synergy. \textit{Uniqueness} represents datasets with no synergy, where all information is contained within a single modality. \textit{Synergy} refers to datasets with complete synergy, where information is distributed across a number of modalities and can only be obtained through cross-modal interactions. \textit{Redundancy} denotes datasets where the same information is present across modalities. \textit{Random} applies to datasets with no meaningful information.  
For each number of modalities (two, three and four), we created a synergy, uniqueness and redundancy dataset as well as a random dataset for the case with two modalities. 

To control for model variability in our experiments, we use the XOR function for the uniqueness and synergy datasets. The XOR function cannot be meaningfully applied to random and redundancy datasets, because it is unclear which information from each modality contributes to the solution, given the multiple possibilities. 
We train fully connected neural networks (FCNNs) on all datasets with three layers: input, hidden (half the input size), and output (half the hidden size). Three fusion strategies are tested: early fusion (data concatenated before input), intermediate fusion (data fused before the hidden layer), and late fusion (data fused before the output layer). Since early fusion captured the most cross-modal interactions, we use it in all subsequent FCNN experiments.
Further details on the dataset generation and experimental setup are provided in Appendix A.The appendices are available in our GitHub repository: \textcolor{RoyalBlue} {{https://github.com/LauraWenderoth/InterSHAP}}.

\subsection{Results}
\textbf{Verification of InterSHAP.} Generally, we anticipate that InterSHAP for the uniqueness dataset will show less cross-modal interaction (expected to be close to 0\%) compared to the synergy dataset (expected to be close to 100\%). 
For the random and redundancy datasets, it is unclear how many cross-modal interactions the FCNN should learn. In the redundancy setting, the FCNN might focus on just one modality or use all modalities since information is distributed across them. In the random setting, there is no inherent information, making the FCNN's learning unpredictable.

Table~\ref{tab:main_results} presents the results of the experiments conducted on the HD-XOR datasets.
InterSHAP predicts the expected amount of cross-modal interaction in the XOR function with 0\% cross-modal interaction for uniqueness and 99.7\% for synergy. In contrast, the FCNN model showed an increase in cross-modal interaction for the uniqueness dataset (0.2\% $\pm$ 0.1) and decrease in synergy (98.0\% $\pm$ 0.5) for the synergy setting. This indicates that the FCNN does not fully capture the systematic structure underlying the dataset's creation with the XOR function. Instead, it sometimes uses both modalities for predictions even when unnecessary, and conversely, may underutilise them when they are required.

InterSHAP indicates that the FCNN attributes approximately 40\% of its behaviour to cross-modal interactions on the redundancy dataset. Especially noteworthy is the outcome on the random dataset, which shows that \mbox{InterSHAP} operates performance-agnostic, as only random performance could be achieved on this dataset. Nevertheless, the model appears to have learned something -- though not relevant to the F1 Score -- evidenced by approximately 60\% of cross-modal interactions.
In summary, InterSHAP demonstrates the expected behaviour and adequately quantifies cross-modal interaction within a model.

\paragraph{Modality Scalability.} 
To test InterSHAP's scalability to more than two modalities, we create HD-XOR datasets with three and four modalities. The InterSHAP values for the FCNN with early fusion across uniqueness, synergy, and redundancy datasets are presented in Table~\ref{tab:InterSHAP HD-XOR 2,3,4 mods}. \looseness -1

\begin{table}[t!]

\centering
\begin{tabular}{lcccccccccc} 
\toprule

  &{Uniqueness} &{Synergy} & Redundancy  \\
  
\midrule
2 Modalities &   0.2  $_{\pm  0.1}$  & 98.0  $_{\pm  0.5}$ & 38.6  $_{\pm  0.5}$   \\
3 Modalities  &  0.6 $_{\pm  0.2}$  & 88.8 $_{\pm  0.5}$ & 51.9  $_{\pm   0.3}$  \\
 4 Modalities  & 1.2 $_{\pm  0.1}$  &64.1 $_{\pm  0.8}$   & 40.2 $_{\pm  0.2}$  \\
\bottomrule
\end{tabular}

\caption[InterSHAP values for FCNN with early fusion on HD-XOR datasets for two, three and four modalities]{InterSHAP values, expressed as percentages, for FCNN with early fusion on HD-XOR datasets with two, three, and four modalities. The results indicate, that \mbox{InterSHAP} works for more than two modalities.}
\label{tab:InterSHAP HD-XOR 2,3,4 mods}
\vspace{-1em}
\end{table}

The expected behaviour is that InterSHAP values for three and four modalities should be the same or similar to those for two modalities, given that the dataset creation process remains unchanged except for the synergy setting.
InterSHAP exhibits the expected behaviour on the uniqueness dataset, with values ranging from \(0.6 \pm 0.2\) to \(2.6 \pm 0.6\). Similarly, values on the redundancy setting show consistent behaviour with larger fluctuations, ranging from a minimum of \(38.6 \pm 0.5\) to a maximum of \(51.9 \pm 0.3\). However, for the synergy setting, InterSHAP indicates a decrease from \(98\%\) cross-modal interactions to \(64\%\) as the number of modalities increases. \looseness -1

\paragraph{Extension to Local Method.}
InterSHAP can be applied to individual samples, as outlined in Section~\ref{Section:Methods local InterSHAP}.
InterSHAP values are averaged over each data point for three runs and standard deviation is calculated, as shown in Table~\ref{tab:main_results}.\looseness -1

On the HD-XOR uniqueness dataset using the XOR baseline function, the local method overestimates the synergy effect, with averages of 1.8\%, compared to the global averages of 0\%. Conversely, for the synergy setting, the local method underestimates the synergy effect, with averages of 96.9\% locally compared to 99.7\% globally. This difference is likely attributable to the masking error source. While this error averages out when considering the entire dataset, it is more pronounced at the data point level.
The difference becomes more apparent when examining the FCNN results.  Particularly noteworthy is the increased standard deviation, which indicates a discrepancy between data points. In contrast to the XOR function, this difference is caused by the neural network and reflects the model's behaviour.

In summary, we have shown that with a controlled model (XOR), cross-modal interactions are slightly overestimated for datasets with near 0\% synergy and slightly underestimated for datasets with near 100\% synergy. For an overall interaction score across the entire dataset, we recommend using the global InterSHAP. However, if the goal is to understand the interaction score at the level of individual data points, local InterSHAP is more appropriate.

\paragraph{Comparison with SOTA.}
To compare InterSHAP with SOTA cross-modal interaction scores, we calculated PID, EMAP and SHAPE, as shown in Table~\ref{tab:main_results}.

\textbf{PID.\,} 
As expected, PID measures fewer cross-modal interactions on the uniqueness compared to the synergy dataset. However, PID is not able to quantify the extent of cross-modal interactions, unlike InterSHAP. In random and redundancy datasets PID also aligns with InterSHAP.  Both show higher values for the random dataset than for the redundancy dataset, indicating that the model learned synergistic relationships.  

\textbf{EMAP.\,}
For clearer comparison, we present EMAP$_{gap}$, calculated as EMAP minus the model's F1 score (see Appendix C,  Table 11), which reflects the extent of cross-modal interactions.
Like other metrics, EMAP also follows the expected pattern, with less synergy observed for the uniqueness dataset than for the synergy dataset. However, EMAP measures no interaction in the model on the random dataset, because it is not performance-agnostic.

\textbf{SHAPE.\,} SHAPE exhibits the expected behaviour on the baseline XOR function for uniqueness and synergy datasets but not for the FCNN model. There are major deviations between the baseline XOR and FCNN results that are not due to model behaviour. While the uniqueness dataset shows similar values for the baseline XOR, the FCNN results differ significantly, with $16.5 \pm 0.06$ for uniqueness.
Additionally, the redundancy dataset results do not align with the previous three metrics. We attribute these discrepancies to the masking and calculation of base values. 

\subsection{Visualisation}
\begin{figure}[t!]
    \centering
    \begin{subfigure}{0.23\textwidth}
        \centering
        \includegraphics[width=\linewidth]{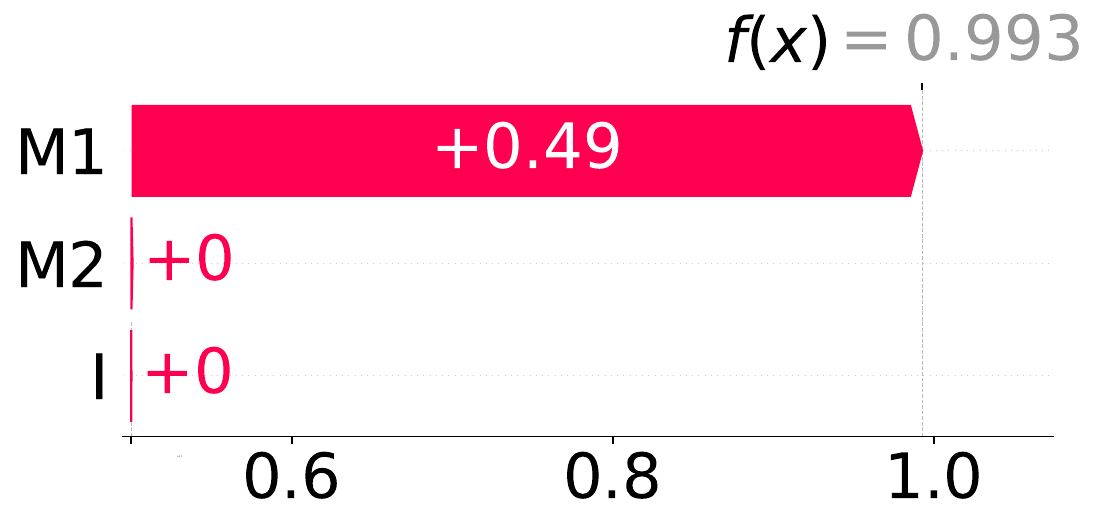}
         \caption{Uniqueness}
    \end{subfigure}
    \begin{subfigure}{0.23\textwidth}
        \centering
        \includegraphics[width=\linewidth]{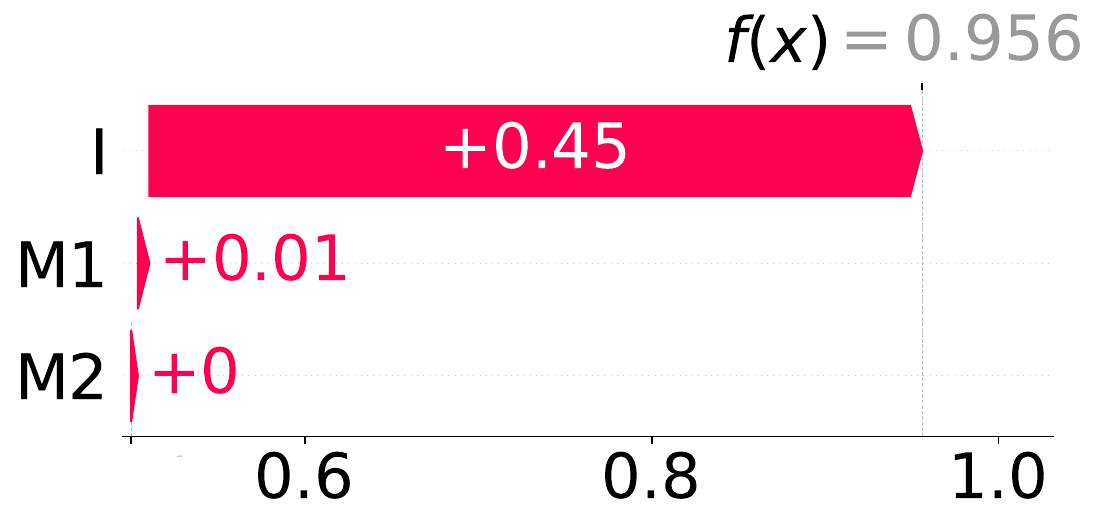}
        \caption{Synergy}
    \end{subfigure}
    \centering
    \begin{subfigure}{0.23\textwidth}
        \centering
        \includegraphics[width=\linewidth]{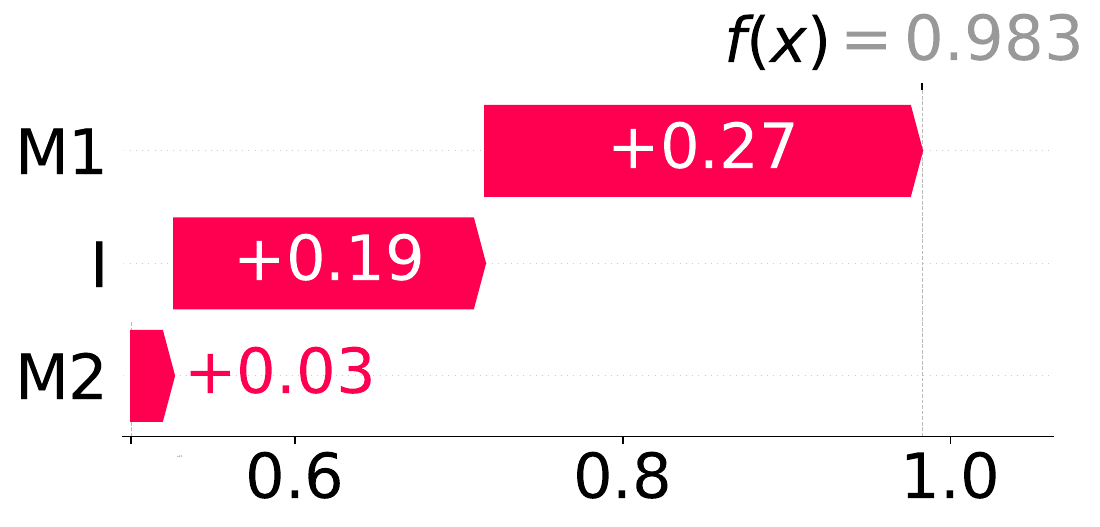}
        \caption{Redundancy}
    \end{subfigure}
    \centering
    \begin{subfigure}{0.23\textwidth}
        \centering
        \includegraphics[width=\linewidth]{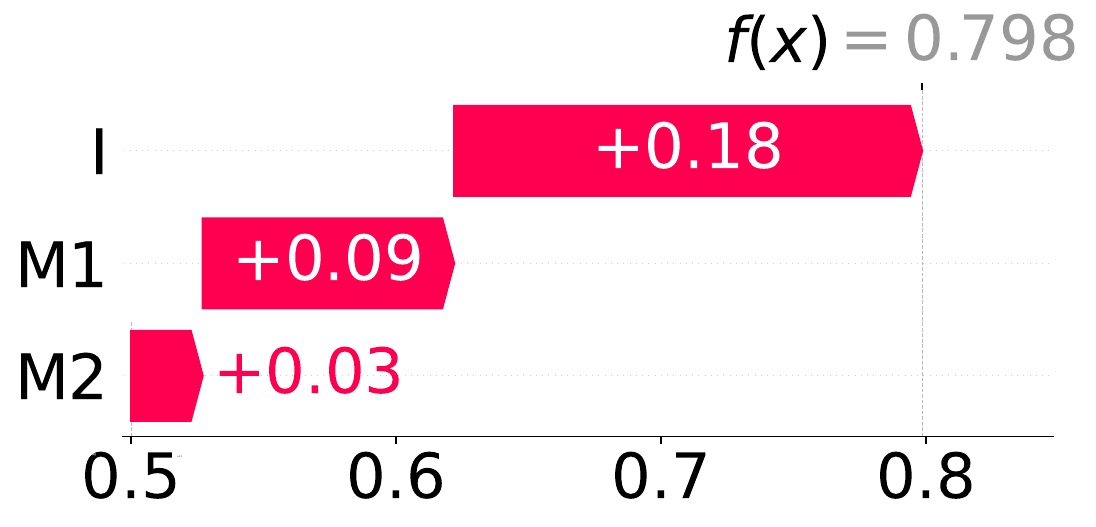}
        \caption{Random}
    \end{subfigure}

    \caption[Visualisation example of InterSHAP]{Visualisation of InterSHAP using the SHAP package integration \cite{lundberg_unified_2017}. The results on the HD-XOR datasets with two modalities for FCNN with early fusion are presented. The x-axes show predicted class probabilities, with a baseline of approximately 0.5 due to the binary classification. M1 represents modality 1, M2 modality 2, and I interactions.}
    \label{fig:Visualitsation early XOR 2 force}
    \vspace{-0.5cm}
\end{figure} \

We designed InterSHAP to be compatible with the visualisation modules of the SHAP implementation \cite{lundberg_unified_2017}. Figure \ref{fig:Visualitsation early XOR 2 force} illustrates the breakdown of model behaviour on the HD-XOR datasets into their components. This representation uniquely shows the relevance of modalities in absolute numbers in addition to cross-modal interactions. For instance, in Figure \ref{fig:Visualitsation early XOR 2 force} (d), it is evident that modality~1 is, on average, more important for the prediction than the interactions. Additionally, the prediction on the random dataset is significantly less certain, with an average probability of only 79.8\%, compared to 95-99\% for the other datasets. \looseness -1

\section{Application to Healthcare Domain}
\label{Chapter:RealWorldResults}
The importance of cross-modal interactions in real-world healthcare datasets is examined by analysing two publicly available multimodal datasets with two modalities with distinct input types.

The first dataset, \textit{multimodal single-cell} \cite{burkhardt_open_2022}, includes RNA and protein data from 7,132 single cells of CD34+ haematopoietic stem and progenitor cells from three donors. Its classification task spans four cell types: neutrophil progenitor (3,121), erythrocyte progenitor (3,551), B-lymphocyte progenitor (97), and monocyte progenitor (363). The data was reduced to 140 proteins and 4,000 genes and split into training (4,993), validation (1,069), and test (1,070) sets (70:15:15 ratio).

The second dataset, \textit{MIMIC-III} \cite{johnson_mimic-iii_2016}, contains anonymised clinical data from ICU patients at Beth Israel Deaconess Medical Center (2001–2012). It includes time series data (12 physiological measurements taken hourly over 24 hours, vector size $12 \times 24$) and static data (5 variables like age, vector size 5). The dataset has 36,212 data points, and is split into training (28,970), validation (3,621), and test (3,621) sets (80:10:10 ratio) \cite{liang_multibench_2021}. The tasks are binary ICD-9 code prediction (group 1) and 6-class mortality prediction. Preprocessing follows \citeauthor{liang_multibench_2021}~\shortcite{liang_multibench_2021}.

{
We use FCNN with all three fusion methods described in Section~\ref{section:setup-experimental} on the multimodal single-cell dataset. For the MIMIC-III dataset, we train two models from MultiBench \cite{liang_multibench_2021}: the MultiBench baseline, which combines an MLP for patient information and a GRU encoder, and the MVAE, which uses a product-of-experts mechanism to combine latent representations from an MLP and a recurrent neural network.}
Table~\ref{tab:single-cell InterSHAP} presents the results concerning the amount of cross-modal interactions for each dataset.

\looseness -1

\begin{table}[t!]
\centering
\begin{tabular}{lrrrrrrr} 
\toprule
&  \multicolumn{3}{c}{Single-Cell} \\
\cmidrule(lr){2-4} 
 &  early & intermediate & late \\
 \midrule
\textbf{InterSHAP} &  1.9 $_{\pm 0.4}$ &1.5 $_{\pm   0.4}$ & 0.4 $_{\pm   0.1}$  \\
PID & 0.08 $_{\pm   0.01}$ & 0.08 $_{\pm  0.01}$ & 0.06 $_{\pm  0.0}$\\
EMAP$_{gab}$ &0 $_{\pm   0}$& 0 $_{\pm   0}$ & 0 $_{\pm  0}$\\
{SHAPE} & 1.0 $_{\pm  0.2 }$ & 0.7 $_{\pm  0.2}$ & 0 $_{\pm  0 }$ \\
\bottomrule
\end{tabular}

\caption[Cross-modal interactions scores on the multimodal single-cell dataset]{Cross-modal interactions scores on the multimodal single-cell dataset for FCNN with early, intermediate and late fusion. InterSHAP aligns with other SOTA methods, capturing the decline in cross-modal information from early to late fusion.}
\label{tab:single-cell InterSHAP}
\end{table}

\subsection{Multimodal Single-Cell}
For the multimodal single-cell dataset, we trained three \mbox{FCNNs} using early, intermediate, and late fusion (performance details in Appendix C), 
and calculated InterSHAP, PID, and EMAP for each model. As shown in Table \ref{tab:single-cell InterSHAP}, InterSHAP values range from 0.4\% to 1.9\%, indicating minimal learned cross-modal interactions. A consistent pattern is observed whereby cross-modal interactions decrease from early to late fusion, which is in accordance with findings from synthetic datasets (see Appendix C.1).

\begin{figure}[t!]
    \centering
    \begin{subfigure}{\linewidth}
        \centering
        \includegraphics[width=\linewidth]{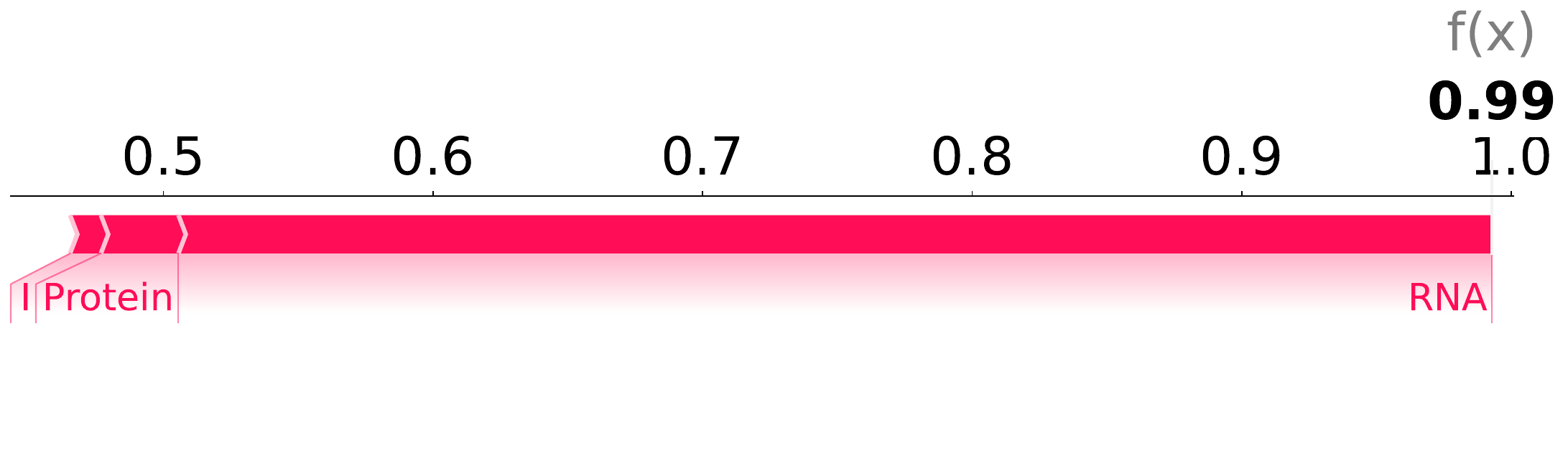}
        \vspace{-3em}
        \caption{Explanation over the whole dataset}
        \vspace{1em}
    \end{subfigure}
    \centering
    \begin{subfigure}{0.49\linewidth}
        \centering
        \includegraphics[width=\linewidth]{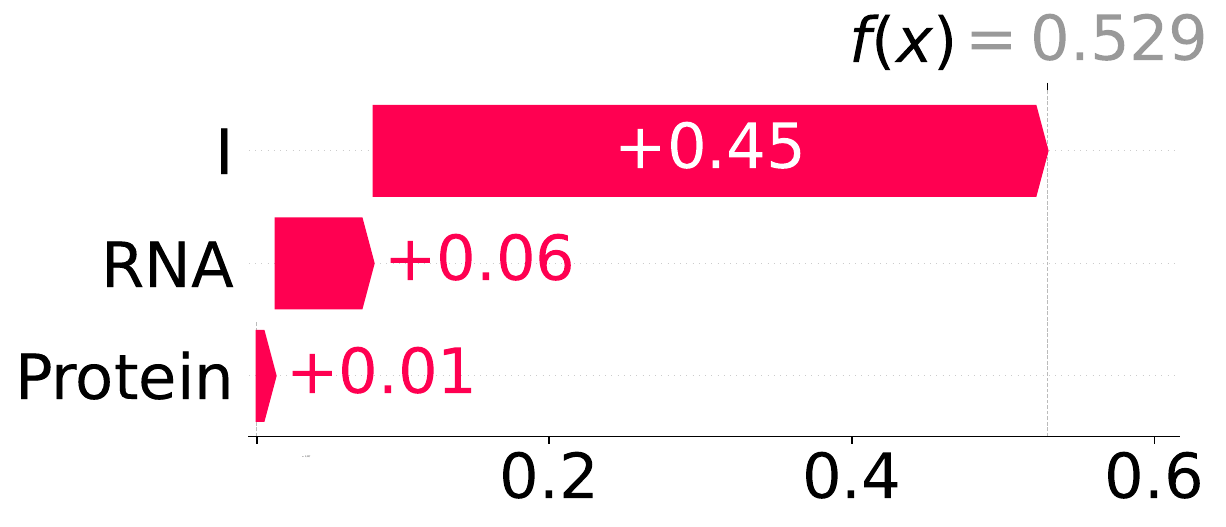}
        \caption{B-Cell Progenitor}
        \vspace{1em}
    \end{subfigure}
    \centering
    \begin{subfigure}{0.49\linewidth}
        \centering
        \includegraphics[width=\linewidth]{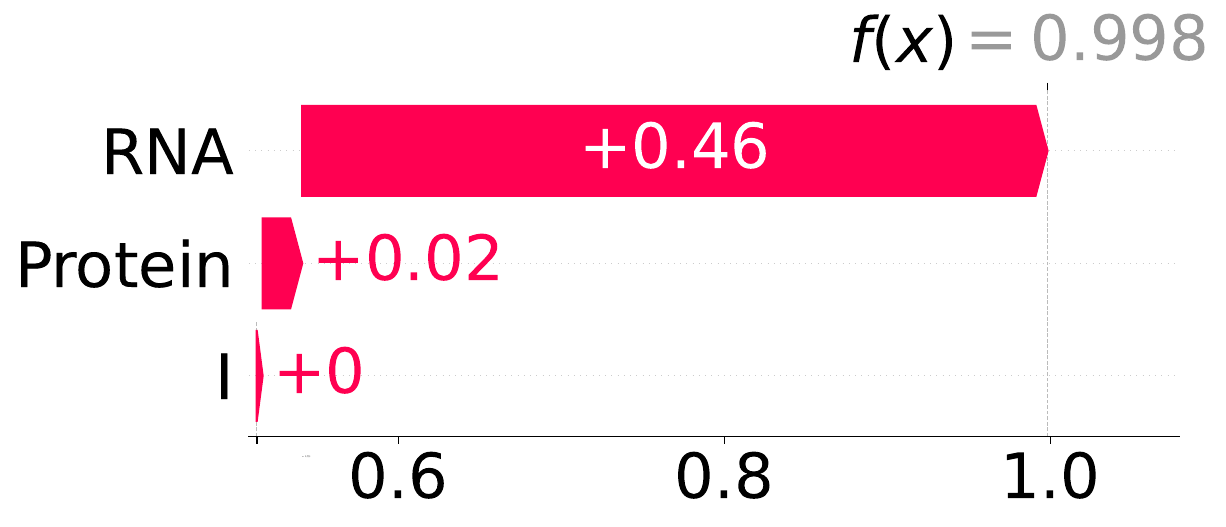}
        \caption{Erythrocyte Progenitor}
        \vspace{1em}
    \end{subfigure}
    \centering
    \begin{subfigure}{0.49\linewidth}
        \centering
        \includegraphics[width=\linewidth]{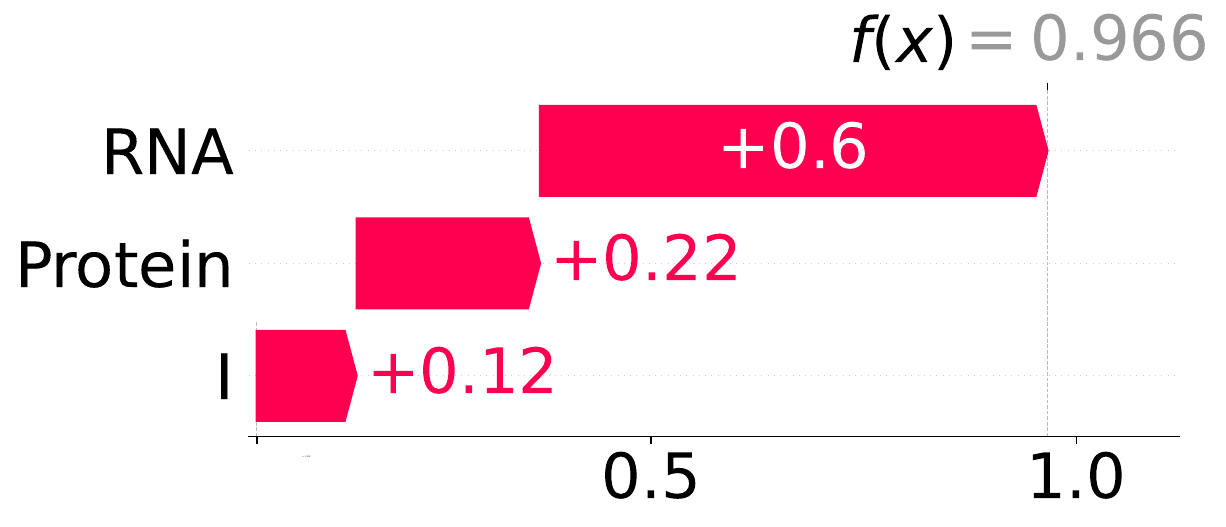}
        \caption{Monocyte Progenitor}
    \end{subfigure}
    \centering
    \begin{subfigure}{0.49\linewidth}
        \centering
        \includegraphics[width=\linewidth]{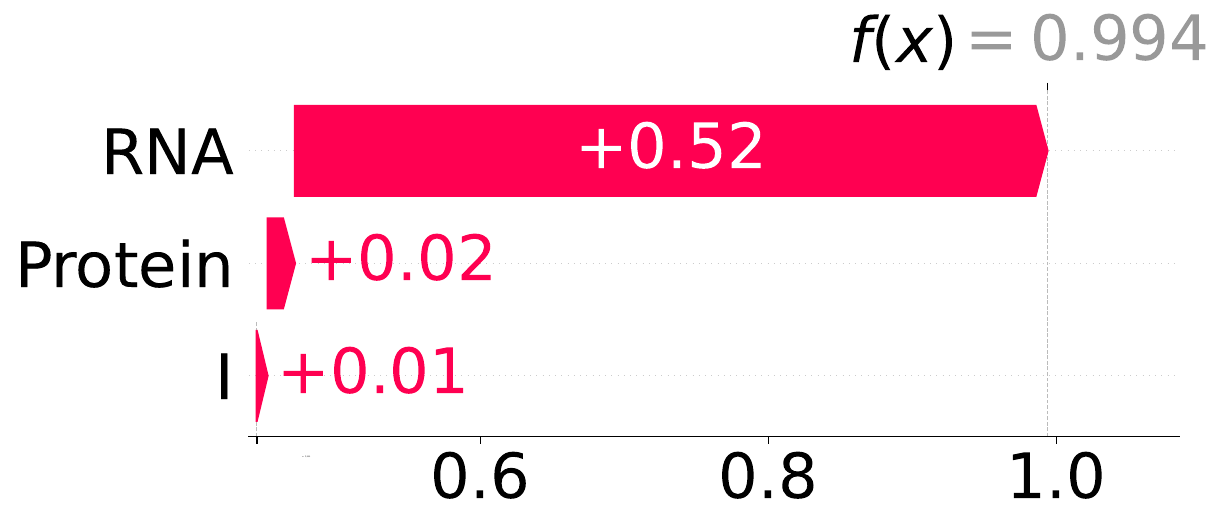}
        \caption{Neutrophil Progenitor}
    \end{subfigure}
    \caption[Visualisations of interactions and modality contributions derived from model trained on the multimodal single-cell dataset]{SHAP visualisations of InterSHAP computed interactions and modality contributions derived from the FCNN early fusion model trained on the multimodal single-cell dataset. {\bf (a)} Force plot of model's behaviour on the whole dataset, with the x-axis representing the class probability of the highest probability class. \mbox{{\bf (b)-(d)}} Breakdown by predicted class.}
    \label{fig:visualtiosation_single-cellInterSHAP}
\end{figure}

We employ SHAP visualisations, as shown in Figure~\ref{fig:visualtiosation_single-cellInterSHAP}, to illustrate modality contributions by predicted class. A clear distinction in measured cross-modal interactions is observed between the larger classes (erythrocytes and neutrophils, comprising 93.5\% of the training dataset) and the smaller classes (monocytes and B-cells). For smaller classes, the model's predictions depend largely on cross-modal interactions, while for larger classes, RNA is the most significant factor influencing the prediction.

The absence of ground truth in real-world datasets is a significant limitation, as the level of synergy or redundancy is unknown. To address this, additional cross-modal interaction scores, PID and EMAP, were computed for comparison with InterSHAP in Table~\ref{tab:single-cell InterSHAP}.
PID displayed behaviour similar to \mbox{InterSHAP} across the three fusion methods, with higher values for early fusion and lower for late fusion. However, PID indicated much higher synergy within the dataset, which is expected since PID focuses on the dataset rather than directly accounting for model behaviour. The authors also note that PID measures consistently exceed those of the dataset \cite[30]{liang_quantifying_2023}. Therefore, to quantify the extent of cross-modal interactions utilised by a model in its predictions, InterSHAP is a more effective solution.
In contrast, EMAP did not detect cross-modal interactions in any models. This difference likely arises because EMAP targets performance rather than directly measuring model behaviour. As depicted in Figure~\ref{fig:visualtiosation_single-cellInterSHAP}, smaller classes benefit significantly from interactions. However, due to their underrepresentation in the dataset, their overall impact on performance remains limited.

We can conclude that models trained on the multimodal single-cell dataset utilise minimal cross-modal interactions to solve the classification task. Exceptions to this are the smaller classes, for which integrating both modalities and concluding information from cross-modal interactions appears more relevant.

\subsection{MIMIC III} 
For the MIMIC dataset, we trained a baseline model and the MVAE model provided in the MultiBench \cite{liang_multibench_2021} (details in Appendix~\ref{app:exp_setup}). 

From the cross-modal interactions presented in Table~\ref{tab:mimic3 InterSHAP}, we observe a consistent pattern: the baseline model shows fewer cross-modal interactions than the more sophisticated MVAE model. \mbox{InterSHAP} values vary between tasks. For deciding if the diagnosis is in ICD-9, lower interactions are measured, ranging between 1.2\% and 6.8\%, compared to the more challenging task of determining mortality, which shows higher interactions around 12\%. 

Comparing InterSHAP with PID and EMAP reveals similar behaviour to the multimodal single-cell dataset, discussed above. PID mirrors the behaviour of InterSHAP but with higher synergy values, while EMAP does not indicate performance-relevant cross-modal scores for the ICD-9 task, but does for the mortality task. This behaviour is expected and aligns well with InterSHAP, which, as it measures performance independently, captures even small amounts of cross-modal interaction, whereas EMAP captures larger performance-relevant ones. 

\begin{table}[t!]
\centering
\begin{adjustbox}{width=\linewidth}
\centering
\begin{tabular}{lrrrrrr} 

\toprule
& \multicolumn{2}{c}{ICD-9} &   \multicolumn{2}{c}{Mortality}\\
\cmidrule(lr){2-3} \cmidrule(lr){4-5} 
&  baseline & MVAE  & baseline & MVAE    \\
 \midrule 
\textbf{InterSHAP}  &  1.2 $_{\pm  0.2}$ &  6.8 $_{\pm  1.3}$ &  11.0  $_{\pm  0.5}$ &   12.3
 $_{\pm  2.8}$ \\
PID & 0.06$_{\pm 0.01}$   & 0.09 $_{\pm  0.01}$  & 0.10 $_{\pm   0.01}$ & 0.11 $_{\pm  0.01}$   \\
EMAP$_{gap}$ & 0  $_{\pm   0}$ & 1.2 $_{\pm  0.0}$  &  $-$0.8 $_{\pm  0.1}$   & 0.9 $_{\pm   0.1}$ \\
{SHAPE} & 0.2 $_{\pm  0}$ & 0.6 $_{\pm  0}$ & 0.2 $_{\pm  0.2}$ & 0.7 $_{\pm  0.2}$\\
\bottomrule
\end{tabular}
\end{adjustbox}

\caption[Cross-modal interactions scores on the MIMIC III dataset]{Cross-modal interactions scores on the MIMIC III dataset for baseline model and MVAE model from MultiBench implementation. InterSHAP aligns with other SOTA methods, capturing greater cross-modal interaction in the baseline compared to MVAE, while uniquely quantifying the proportional contribution of cross-modal interactions.}
\label{tab:mimic3 InterSHAP}
\end{table}

\section{Discussion and Future Work}
InterSHAP quantifies interaction strength as an interpretable percentage and works effectively with unlabelled data. Although the value of a performance-agnostic metric may be debated, InterSHAP is particularly useful for model development and debugging. For example, in our imbalanced single-cell study, smaller classes with minimal impact on the overall performance exhibited a higher reliance on cross-modal interactions, highlighting areas where modality integration could be improved. By capturing all cross-modal interactions -- not just those linked to performance -- \mbox{InterSHAP} provides deeper insights into low-performing models and can conclusively identify the absence of interactions, avoiding spurious results seen with performance-dependent methods.

Despite these benefits, we note some limitations of InterSHAP.
The first is the exponential increase in runtime with the number of modalities, as all possible coalitions must be computed due to the lack of general approximation methods for SII. However, this is less critical in practice, as models are typically trained on fewer than ten modalities \cite{barua_systematic_2023,xu_comprehensive_2024}. Future work could explore approximation methods to address this computational challenge. \looseness -1

The second limitation is that we could not clearly demonstrate that InterSHAP functions as intended for synergy datasets with more than two modalities due to two factors: (1)~The synergy in the training dataset decreases as the number of modalities increases. This phenomenon is attributed to the definition of XOR for more than two modalities: only one occurrence of 1 will result in the label being true, with the rest being 0. Consequently, if a 1 is present in two modalities, the label (i.e., false) is already known. (2)~The F1 score decreases: 94.4 \(\pm\) 0.3 for two modalities, 79.2 \(\pm\) 0.2 for three modalities, and 76.4 \(\pm\) 0.2 for four modalities. This indicates that the synergy present in the dataset is not fully learned, leading to the model likely relying less on cross-modal interactions for prediction. The observed decrease is likely attributable to the two factors, although the potential contribution of \mbox{InterSHAP} cannot be ruled out. Although \mbox{InterSHAP} relies on the Shapley Interaction Index (SII) and captures pairwise modality interactions, it may not fully address complex cross-modality interactions. 
Future research could explore alternative interaction measures than the SII.  Other unimodal indices at the feature level could be also applied to quantify cross-modal interactions in datasets with more than two modalities to overcome this limitation. Potential alternatives for evaluation include Faith-SHAP \cite{tsai_faith-shap_2023} and the Shapley-Taylor Interaction Index \cite{sundararajan_shapley_2020}.

\section{Conclusion} 
We introduce InterSHAP, a multimodal explainability metric designed to quantify cross-modal interactions and modality contributions. 
InterSHAP provides a straightforward interpretation of cross-modal interactions, expressed as a percentage of overall model behaviour, which can be applied to multimodal learning tasks with any number of modalities. Besides explainability, it allows for investigating the capabilities of different model architectures in how they exploit multimodal interactions across real-world datasets. As such, InterSHAP overcomes the limitations of existing methods, akin to constraints to only two modalities and the lack of reliable, performance-independent methods. Our experiments on synthetic and real-world data demonstrate the ability of InterSHAP to accurately measure the presence (or absence) of cross-modal interactions, while allowing for global (dataset-level) and local (sample-level) explanations.   

\newpage
\section*{Acknowledgments}
LW acknowledges support from the Hans-Böckler-Foundation. KH acknowledges support from the Gates Cambridge Trust via the Gates Cambridge Scholarship. NS and MJ acknowledge the support of the U.S. Army Medical Research and Development Command of the Department of Defense; through the FY22 Breast Cancer Research Program of the Congressionally Directed Medical Research Programs, Clinical Research Extension Award GRANT13769713. Opinions, interpretations, conclusions, and recommendations are those of the authors and are not necessarily endorsed by the Department of Defense. \looseness -1
\bibliography{aaai25}
\clearpage
\appendix
\section{Synthetic Data Generation}
\label{Section:synthetic Datasets methods}
Since synthetic dataset generation presents inherent challenges, including data distribution bias, insufficient noise levels, over-smoothing, and inconsistencies \cite{hao_synthetic_2024}, we draw inspiration from two existing approaches to create the synthetic datasets \cite{liang_quantifying_2023,hessel_does_2020} to formulate our synthetic data generation strategy HD-XOR. Additionally, we use a second synthetic data generation method called SUM. For each generation method, there exists an ensemble of datasets displayed in Table \ref{tab:overview_synthic_datasets_ensambles}. An ensemble refers to a set of datasets created using the same method (HD-XOR or SUM) with varying degrees of synergy, from setting uniqueness (no synergy) to setting synergy (full synergy). In the following, we will first present our strategy (HD-XOR) for creating the dataset ensemble. Subsequently, we will briefly introduce the strategy (SUM dataset) described by \citeauthor{liang_quantifying_2023}~\shortcite{liang_quantifying_2023}.

\paragraph{High-Dimensional XOR Dataset.} 
\begin{algorithm}[ht!]
\caption{Generate synthetic data}\label{algo:gen_Data}
\begin{algorithmic}[1]
\Procedure{data\_distribution}{setting, $N$: number of samples, $M$: number of modalities, dims $\in \mathbb{Z}^M,$}
    \State $\iota \gets 2$
    \If{$M \geq 3 \quad \land \quad$setting is synergy}
        \State $\iota \gets M$
    \EndIf
    \State $I,y \gets \{i_1^i,\dots,i_M^i\}_{i=1}^N, \{y^i\}_{i=1}^N \quad \{i_1^i,\dots,i_M^i\}, y^i = \text{GEN\_INFO}(A=\iota)$
    \State
    \If{setting is redundancy}
        \State $M_m \in \mathbb{R}^{\text{dims}_m \times 2 \cdot d} \sim \mathcal{U}(-0.5, 0.5), \quad \forall i \in \{1,\dots,M\}$
        \State $x^i \gets [i_1^i, i_2^i]$, \quad $\forall i \in \{1,\dots,N\}$
        \State $X \gets \{x_1^i,\dots,x_M^i\}_{i=1}^N$
    \ElsIf{setting is uniqueness}
        \State unique modality $u \in \{1,\dots,M\}$
        \State $M_m \in \mathbb{R}^{\text{dims}_m \times 2 \cdot d} \sim \mathcal{U}(-0.5, 0.5), \quad \forall i \in \{1,\dots,M\}$
        \State $x^i \gets [i_1^i, i_2^i]$, \quad $\forall i \in \{1,\dots,N\}$
        \State  $x_j \in \mathbb{R}^{2 \cdot d} \sim \mathcal{N}(0,1), \quad \forall j \{1,\dots,M\}, j \neq u$
        \State $X \gets \{x_1^i,\dots,x_u^i,\dots,x_M^i\}_{i=1}^N$
    \ElsIf{setting is synergy}
        \State $M_m \in \mathbb{R}^{\text{dims}_m \times d} \sim \mathcal{U}(-0.5, 0.5), \quad \forall i \in \{1,\dots,M\}$
        \State $X \gets I$
    \EndIf
    \State  $X \gets \{M_1 \cdot x_1^i,\dots,M_M \cdot x_M^i\}_{i=1}^N$
    \State \textbf{return} $X,y$
\EndProcedure
\vspace{0.5cm}
\end{algorithmic}

\begin{algorithmic}[1]
\Procedure{gen\_info}{$A$: number of information dimensions}
    \State  $i \gets \left\{i_a \in \{0,1\} \sim \text{Bernoulli}(\frac{1}{2})\right\}_{a=1}^A $
    \State $y \gets i_1 \oplus i_2 \oplus \dots \oplus i_A = \bigoplus_{a=1}^{A} i_a$
    \State Project $i$ in higher dimensional space:
    \For{$a \gets 1$ to $A$}
        \If{$i_a$ is 0} 
            \State $i_a \gets i_a \in \mathbb{R}^d \sim \mathcal{N}(-0.35,1), \quad a \in \{1, \dots, A\}$
        \Else 
             \State $i_a \gets i_a \in \mathbb{R}^d \sim \mathcal{N}(0.35,1), \quad a \in \{1, \dots, A\}$
        \EndIf
    \EndFor
    \State \textbf{return} $i,y$
    \EndProcedure
\end{algorithmic}

\end{algorithm}
The first synthetic dataset generation strategy is the high-dimensional XOR (HD-XOR). This method builds on the XOR function and generates high-dimensional datasets with any number of modalities and dimensions. To construct the HD-XOR dataset, we drew inspiration from methodologies proposed by \citeauthor{hessel_does_2020}~\shortcite{hessel_does_2020} and \citeauthor{liang_quantifying_2023}~\shortcite{liang_quantifying_2023}. 

The procedure to generate datasets using  HD-XOR is described in Algorithm \ref{algo:gen_Data} in detail. Firstly,  for each sample in the dataset, two numbers, denoted by \(i_1\) and \(i_2\), are randomly drawn from a fair Bernoulli distribution representing binary features. Suppose the synergy setting involves more than two modalities, as many numbers as modalities are drawn (line $2-5$). The label for each data point is then calculated using the XOR operation. Subsequently, \(i_1\), \(i_2, \dots i_M\), are projected into a higher dimensional space using a normal distribution as described in \cite{hessel_does_2020}. To obtain vectors with the information encoded by the binary integers, vectors from normal distributions with different mean values (negative for $0$, positive for $1$).
\begin{table*}[t!]
\centering

\begin{tabular}{ccccc}
\toprule
\multicolumn{2}{c}{\textbf{2 Modalities}} & \textbf{3 Modalities} & \textbf{4 Modalities} \\
\cmidrule(lr){1-2}  \cmidrule(lr){3-3}  \cmidrule(lr){4-4}
HD-XOR & SUM & HD-XOR & HD-XOR \\
\midrule
\begin{tabular}[t]{@{\hspace{0.5em}}lr@{\hspace{0.5em}}}Uniqueness  
&\textcolor{WildStrawberry}{$\downarrow$} \\ Synergy &\textcolor{Green}{$\uparrow$}\\ Redundancy & ? \\ Random & ?\end{tabular} & \begin{tabular}[t]{@{\hspace{0.5em}}lr@{\hspace{0.5em}}}Uniqueness 1 
&\textcolor{WildStrawberry}{$\downarrow$} \\ Uniqueness 2 
&\textcolor{WildStrawberry}{$\downarrow$} \\ Mix 1 &\textcolor{Goldenrod}{$\leftrightarrow$}\\ Mix 2 &\textcolor{Goldenrod}{$\leftrightarrow$}\\ Synergy&\textcolor{Green}{$\uparrow$}\\ Redundancy& ?\end{tabular} &  \begin{tabular}[t]{@{\hspace{0.5em}}lr@{\hspace{0.5em}}}Uniqueness  
&\textcolor{WildStrawberry}{$\downarrow$} \\ Synergy&\textcolor{Green}{$\uparrow$}\\ Redundancy &?\end{tabular} & \begin{tabular}[t]{@{\hspace{0.5em}}lr@{\hspace{0.5em}}}Uniqueness 
&\textcolor{WildStrawberry}{$\downarrow$}\\ Synergy &\textcolor{Green}{$\uparrow$}\\ Redundancy & ?\end{tabular}\\
\bottomrule
\end{tabular}
\caption[Overview of generated synthetic datasets]{Overview of generated synthetic datasets using two different generation methods, HD-XOR and SUM. The arrows indicate the degree of synergy in the dataset. A red arrow (\textcolor{WildStrawberry}{$\downarrow$}) signifies no synergy, a green arrow (\textcolor{Green}{$\uparrow$}) indicates maximum synergy, and a yellow arrow (\textcolor{Goldenrod}{$\leftrightarrow$}) represents moderate synergy. The question marks (?) denote uncertainty about the amount of synergy in the dataset, as it was created with maximum redundancy.}
\label{tab:overview_synthic_datasets_ensambles}
\end{table*}

Subsequently, the information of the vectors must be distributed for the three distinct settings (redundancy, synergy, and uniqueness), employing a methodology analogous to that described in \cite{liang_quantifying_2023}. 
In the redundancy setting, the variables \(i_1\) and \(i_2\) are concatenated and then copied \( M \) times  (where \( M \) represents the desired number of modalities) for each sample. In the uniqueness setting, the components \(i_1\) and \(i_2\) are also concatenated but only written to the desired modality. All other modalities are filled with vectors drawn from a normal distribution with a mean value of 0. Finally, in the synergy setting, the components \(i_1\), \(i_2, \dots i_M\) are distributed according to the modalities $1, 2, \dots M$, respectively.
To generate the desired dimensions, we draw $M$ matrices (one for each modality) from a uniform distribution. The matrix of each modality is multiplied element-wise with the corresponding normalised vector of unit length belonging to the same modality. This ensures that each vector in each modality is unique in the redundancy setting. The labels remain unchanged during the generation of the modalities.
Additionally, we created a random dataset to evaluate behaviour on data where nothing meaningful can be learned. The vectors for the modalities are randomly generated with the desired dimensions and a mean of 0, while the labels are randomly selected from 0 and 1.

The HD-XOR dataset faces limitations when applied to more than two modalities in the synergy setting. For more than two modalities, the dataset no longer contains maximum synergy because XOR is defined as the sum of all integers to be one: \( \sum_{a=1}^M i_a = 1 \). When there are already two positive values in the first modalities, the label is known to be negative. Thus, we know that synergy decreases with an increasing number of modalities.

\paragraph{SUM Dataset.} 
\label{Sec:PID_dataset}
Additionally, we employ a second strategy to generate synthetic datasets for two modalities from \citeauthor{liang_quantifying_2023}~\shortcite{liang_quantifying_2023} to validate the functionality of InterSHAP across a diverse range of datasets. Like the HD-XOR dataset, the SUM dataset consists of an ensemble of datasets with varying levels of synergy. In addition to the three previously described settings — synergy, redundancy, and uniqueness — this generation approach allows for creating mixed datasets containing a mixture of unique and synergistic information.

To create the SUM datasets, we draw three vectors, \(x_1\), \(x_2\), and \(r\), from a normal distribution and normalise them. We then concatenate the vectors \(x_1\) and \(r\) as well as \(x_2\) and \(r\) to form a vector for each modality, containing modality-specific information (denoted by \(x\)) and shared information (denoted by \(r\)). 
We use different parts of the vector to calculate the labels depending on the desired setting (uniqueness, synergy, redundancy, mix). Regarding redundancy, we consider only the component \(r\) to ensure information redundancy across both modalities. For uniqueness, we use either the subcomponent \(x_1\) or \(x_2\). For synergy, both subcomponents \(x_1\) and \(x_2\) contribute to the calculation of the label. For the mix setting, we combine parts of \(x_1\) and \(x_2\). We use a sum function to calculate the labels, hence the name. For binary labels, the label is assigned 1 if the sum is greater than 0.5, otherwise 0. 

The SUM dataset is limited in the synergy setting because the sum of one modality alone can exceed or be very close to 0.5, making the second vector unnecessary or resulting in guesses that deviate from the expected 50\% probability in binary classification.

\section{Experimental Setup}
\label{app:exp_setup}

\subsection{Synthetic Datasets} We created synthetic datasets with varaying degrees of synergy using two distinct methods HD-XOR and SUM described in Section Synthetic Data Generation. For each generation method, there exists an ensemble of datasets displayed in Table \ref{tab:overview_synthic_datasets_ensambles}. An ensemble refers to a set of datasets created using the same method (HD-XOR or SUM) with varying degrees of synergy, from setting uniqueness (no synergy) to setting synergy (full synergy). 

The generated datasets consist of 20,000 samples each, with specific dimensionality configurations customised for individual modalities.  The two-modal configuration of the HD-XOR datasets comprise dimensions of 110 and 90, while the SUM datasets feature dimensions of 100 for both modalities. For the HD-XOR datasets incorporating three, four, and five modalities, the dimensions are adjusted accordingly: (110, 90, 100), (110, 90, 100, 110), and (110, 90, 100, 110, 90).  Each dataset is partitioned into three subsets - validation, testing, and training - with 15\% of the samples allocated to the validation and testsets and the remaining 70\% designated for training purposes. Moreover, the number of output classes remains fixed at two for both datasets.

\subsection{Healthcare Datasets}
\label{Section:real-world datasets description}
We use two publicly available multimodal datasets. The first is called multimodal single-cell dataset and includes surface proteins and RNA sequencing modalities from hematopoietic stem cells. The second dataset is the MIMIC III dataset, which contains clinical data from intensive care units. This dataset comprises general patient data and time-series data from sensors used in intensive care. 

\paragraph{Mulitmodal Single-Cell.} 

The Multimodal Single-Cell dataset, constructed by \citeauthor{burkhardt_open_2022}~\shortcite{burkhardt_open_2022}, integrates multi-omics data comprising gene expression profiles (RNA) and protein levels (protein) of single cells derived from mobilised peripheral CD34+ haematopoietic stem and progenitor cells isolated from three healthy human donors. Each cell within the dataset is assigned a cell type, including mast cell progenitor, megakaryocyte progenitor, neutrophil progenitor, monocyte progenitor, erythrocyte progenitor, haematopoietic stem cell, and B-cell progenitor.

\begin{table}
\centering

\begin{tabular}{ccccccc} 
\toprule
\multicolumn{4}{c}{Class Distribution}  \\
\cmidrule(lr){1-4} 
 Neutrophil & Erythrocyte & B-Lymphocytes & Monocyte \\
\midrule
  43.7\% & 49.8\% & 1.4\% & 5.1\%  \\
\bottomrule
\end{tabular}
\label{tab:single cell_overview_classes_Samples}
\caption{Details of Single Cell.}
\end{table} 

\textbf{Preprocessing.\,} To facilitate analysis, a subset of four cell classes was selected, comprising two physiologically frequent cell classes, neutrophil progenitor and erythrocyte progenitor, which are overrepresented in the dataset, and two rarer cell classes, B-lymphocytes progenitor and monocyte progenitor. This resulted in 7132 data points, with a class distribution of 3121, 3551, 97, and 363, respectively. Furthermore, the surface protein modality was randomly reduced to 140 proteins, and the RNA modality was reduced to 4000 genes. Furthermore, the dataset is organised into training, validation, and test sets following a 70:15:15 split, respectively,  yielding 4,993 training data points, 1,069 validation data points, and 1,070 test data points.

\paragraph{MIMIC III.}

The Medical Information Mart for Intensive Care III (MIMIC III) dataset is a comprehensive collection of deidentified clinical data from patients treated in intensive care units at Beth Israel Deaconess Medical Center in Boston, Massachusetts, USA between 2001 and 2012 \cite{johnson_mimic-iii_2016}. 

\begin{table}[ht!]
\centering

\begin{adjustbox}{width=\linewidth}
\begin{tabular}{cccccccc} 
\toprule
\multicolumn{8}{c}{Class Distribution}  \\
\cmidrule(lr){1-8} 
\multicolumn{2}{c}{ICD} &   \multicolumn{6}{c}{Mortality}  \\
\cmidrule(lr){1-2}  \cmidrule(lr){3-8}
 No & Yes  & 1d & 2d & 3d & 7d & 1 year & $>$ 1 year \\
\midrule
82.5\% & 17.5\% &  76.0\% & 0.4\% & 1.3\% & 1.0\% & 11.0\% & 10.3\% \\

\bottomrule
\end{tabular}
\end{adjustbox}

\caption{Details of MIMIC III, ICD 1 and mortality.}
\label{tab:MIMIC_overview_classes_Samples}
\end{table}
\textbf{Preprocessing.\,}
The dataset was preprocessed by \citeauthor{liang_multibench_2021}~\shortcite{liang_multibench_2021}, who followed the preprocessing protocol of \citeauthor{purushotham_benchmarking_2018}~\shortcite{purushotham_benchmarking_2018} and reported that the process took approximately one to two weeks. The original dataset contained mistakes due to noise, missing values, outliers, duplicate or incorrect records, and clerical errors, as detailed in \cite{purushotham_benchmarking_2018}. 

The preprocessed dataset comprises two modalities: time series data and static information on patients. The time series data comprises 12 time-stamped, nurse-verified physiological measurements, including heart rate, arterial blood pressure, fluid balance, and respiratory rate, taken every hour over a 24-hour period, resulting in a vector of size $12 \times 24$. In contrast, the static data comprises five medical information variables, including the e.g patient's age, represented in a vector of size five.

The available tasks comprise two ICD-9 code predictions and a mortality prediction. The ICD-9 code predictions entail a binary classification to determine whether a patient fits any ICD-9 code within group 1 (140-239) or group 7 (460-519). The International Classification of Diseases (ICD) is a widely recognised classification system for medical diagnoses published by the World Health Organisation (WHO). It provides a standardised framework for coding and categorising diseases, allowing for accurate and consistent reporting of health information \cite{organization_international_1978}. We only used ICD-9 group 1 for our experiments. The mortality prediction, on the other hand,
is a six-class prediction to determine whether the patient dies within one day, two days, three days, one week, one year, or longer than one year. 

Following the split proposed by \citeauthor{liang_multibench_2021}~\shortcite{liang_multibench_2021}, the data was randomly divided into three sets: a training set, a validation set, and a test set with a ratio of 80:10:10, respectively. This resulted in 28,970 training data points, 3,621 validation data points, and 3,621 test data points, for a total of 36,212 data points. Please refer to Table \ref{tab:MIMIC_overview_classes_Samples} for further details.

\textbf{Access Restrictions and Privacy.\,} The data has been deidentified in accordance with the relevant access restrictions and privacy considerations. This process involved removing identifying data elements such as telephone numbers and addresses. For further details, please refer to \citeauthor{johnson_mimic-iii_2016}~\shortcite{johnson_mimic-iii_2016}. Furthermore, all researchers working with this dataset must have completed courses in the protection of human research participants, including those pertaining to HIPAA (Health Insurance Portability and Accountability Act) requirements.

\subsection{Models}
The model is one of the three key components influencing the outcome of cross-modal interaction. Therefore, we use two models: a baseline based on the XOR function and a neural network. This approach allows us to control the model behaviour with the baseline, ensuring that the only uncontrolled component is the cross-modal interaction metric. 

\paragraph{XOR Function.} 
The baseline refers to the function underlying the creation of the HD-XOR dataset. We utilise the XOR function for the controlled setting because it consistently produces the correct output, ensuring that the model reflects the maximum synergy in the dataset. The baseline model $f$ is described in Algorithm \ref{alg:function_f}. Here, \(i = [i_1, i_2, \ldots, i_M]\) represents the different modalities, while $[1.0 , 0.0]$ and $[0.0 , 1.0]$ denote the class-wise probabilities. The function get\_unique(setting) in line 3 determines which modality contains the information and returns the corresponding integer.

\begin{algorithm}[ht]
\caption{Baseline XOR Function $f$}\label{alg:function_f}
\begin{algorithmic}[1]
\Procedure{XOR}{$i = [i_1, i_2, \ldots, i_M]$, setting}
    \If{setting $==$ uniqueness}
        \State unique\_modality $\gets$ get\_unique(setting)
        \State $i \gets i[\text{unique\_modality}]$
    \EndIf
    \If{$\sum_{a=1}^{M} i[a] == 1$}
        \State \textbf{return} $[1.0, 0.0]$
    \Else
        \State \textbf{return} $[0.0, 1.0]$
    \EndIf
\EndProcedure
\end{algorithmic}
\end{algorithm}

\begin{table*}[ht!]
\centering
\begin{tabular}{lcccccccccc} 
\toprule
  &{{Uniqueness 1}} &{{Uniqueness 2}} & Mix 1 & Mix 2 & {Synergy}  &{{Redundancy}}\\

\midrule
InterSHAP &   2.6  $\pm$ 0.6 & 2.3 $\pm$ 0.3  & 9.7 $\pm$ 0.9 & 7.1 $\pm$ 0.5 & 15.1 $\pm$  0.5 &  10.5 $\pm$ 0.2 \\
${\text{InterSHAP}_{local}}$  &  9.3 $\pm$ 10.2 & 9.0 $\pm$ 12.7 & 14.7 $\pm$ 9.9 & 13.8 $\pm$ 9.8 & 22.1 $\pm$ 11.2 & 22.3 $\pm$ 13.3 \\
\bottomrule
\end{tabular}

\caption[InterSHAP values for FCNN with early fusion on SUM datasets]{InterSHAP values are depicted as percentages for the FCNN with early fusion on SUM datasets. InterSHAP measures the fewest cross-modal interactions in uniqueness datasets and the most in synergy datasets. Mixed datasets, as expected, fall between synergy and uniqueness, indicating that InterSHAP accurately quantifies gradations of cross-modal interactions.}

\label{tab:Results_InterSHAP_SUM_earlyfusion}
\end{table*}

It is important to note that the fundamental XOR function only applies to synergy and all uniqueness datasets. This limitation arises due to the necessity of determining which information from the modalities is included in the calculation. For the redundancy setting, however, it is unclear whether a modality can be entirely disregarded or only a portion. Moreover, the XOR function does not apply to random datasets, as it was not the basis for their creation.

\paragraph{FCNN.} \label{Sec:Fccn} 
For the synthetic datasets, fully connected neural networks (FCNNs) comprising three layers are employed. In an FCNN, each neuron in one layer is connected to every neuron in the adjacent layer. The hidden dimensions are half the input dimensions, and the output dimensions are half the hidden dimensions. Three different fusion strategies are examined: early, intermediate, and late fusion. Early fusion involves concatenating or fusing the data before inputting it into the network. Intermediate fusion occurs before the hidden layer, while late fusion occurs before the output layer.

 We experimentally validate InterSHAP starting with two modalities, followed by scalability to more than two modalities, and then extending to the local method, with subsequent visualisation options available due to integration with the SHAP package \cite{lundberg_unified_2017}. Finally, we will compare InterSHAP with existing cross-modal interaction scores on synthetic datasets.

\paragraph{MultiBench Baseline.} 
We use the baseline provided by the MultiBench implementation  \cite{liang_multibench_2021}. The baseline combines a multi-layer perception (MLP) for patient information and a gated recurrent unit (GRU) encoder to process sequential data physiological measurements. Their outputs are concatenated in a fusion layer, forming a joint representation for classification through an MLP classification head with two hidden layers.

\paragraph{MVAE.} The MMultimodalVariational Autoencoder (MVAE) \cite{wu_multimodal_2018} is structured similarly to the baseline but employs a different fusion mechanism. For our implementation, we utilise the MultiBench implementation \cite{liang_multibench_2021} without any alterations. The encoder consists of an MLP for patient information and a time series encoder for physiological measurements (recurrent neural network), which maps to a 200-dimensional vector in the latent space. The fusion involves using a product-of-experts mechanism to combine these latent representations, which are subsequently utilised for classification using an MLP.

\subsection{Implementation}   We develop an open-source implementation of \mbox{InterSHAP}, availble at {\color{RoyalBlue} \url{github.com/LauraWenderoth/InterSHAP}}. This implementation includes InterSHAP alongside other methods such as PID, EMAP, and SHAPE. Additionally, our implementation seamlessly integrates with the widely-used SHAP visualisation package \cite{lundberg_unified_2017}. We employed a masking strategy by replacing the modalities of one data point with those from another data point drawn from the training data. This process was repeated \( n \) times, and the results were averaged.
For the target, we chose the probabilities of classification, specifically the highest class probability, because it ensures performance independence and represents the decisive outcome. 

For PID, we use the CVX estimator and report the synergy value as the cross-modal interaction score. The macro F1 score is chosen as the target metric for both EMAP and SHAPE, and they are computed without further modifications. EMAP was calculated using an open-source implementation, whereas SHAPE required re-implementation based on the original paper, assuming the model always predicts the majority class to establish its baseline performance.

\textbf{Training Details.\,} All training runs with synthetic data were conducted using three seeds: 1, 42, and 113. The learning rate was set at $1e-4$ without early stopping, and the optimiser used was Adam with a weight decay of $1e-4$.
Models with two and three modalities were trained for 200 epochs, while models with four were trained for 250 epochs.

 \section{Additional Results} 
 \label{app:additionalresults}
 \subsection{Synthetic Data}

 \paragraph{Consistency of InterSHAP.}

We tested InterSHAP on another data generation strategy to ensure its consistent, correct performance. Table \ref{tab:Results_InterSHAP_SUM_earlyfusion} displays the outcomes of InterSHAP conducted on FCNNs with early fusion trained on each SUM dataset. The results demonstrate that InterSHAP effectively quantifies cross-modal interactions as intended, with minimal cross-modal interactions detected for uniqueness settings ($2.6 \pm 0.6$ for uniqueness 1 and $2.3 \pm 0.3$ for uniqueness 2), maximal for the synergy setting ($10.5 \pm 0.2$), and falling in between for mixed synergy settings ($9.7 \pm 0.5$ for mix 1 and $7.1 \pm 0.5$ for mix 2). This aligns with the anticipated behaviour.

Compared to the XOR dataset, the cross-modal interactions computed by InterSHAP are generally smaller. Even in the synergy setting, only 15\% of the model behaviour is attributed to cross-modal interactions. This behaviour is supported by the F1 scores, which indicate that 79\% can be achieved with a single modality on average while using both modalities yields 87.6\% on average. Therefore, it can be inferred that the dataset inherently contains less synergy than the XOR dataset.

Examining the results for InterSHAP$_local$, on average, higher synergy is estimated, albeit with a higher standard deviation. However, it is also evident here that InterSHAP$_local$ estimates less synergy for both unique settings than the mixed settings and, on average, the most for the synergy setting.
In conclusion, InterSHAP is a reliable metric for cross-modal interaction for two modalities.

\paragraph{Effect of Fusion Method on Interactions.} 
\label{app:effectfusionmethod}

To determine the most effective fusion method, we train the FCNN on HD-XOR datasets with two modalities using early, intermediate, and late fusion, and then compute InterSHAP for all three fusion methods. The results are presented in Table \ref{tab:HD-XOR fusion stratagies}. 

All fusion approaches measure approximately the same amount of synergy on the unique dataset. However, this differs in the synergy datset. There, early fusion exhibits the highest proportion of synergy effects. Nevertheless, it is noteworthy to mention that the performance of intermediate fusion is lower ($80.0 \pm 1.5$) compared to early fusion ($92.4 \pm 0.3$), indicating that this effect may be attributed to the lower performance. The results are particularly pronounced for late fusion ($49.7 \pm 1.2$), where InterSHAP could not detect cross-modal interactions.
Interestingly, although the F1 score has decreased for intermediate fusion in the redundancy setting, the cross-modal interactions increase. This suggests a redundant additive interaction effect, where only specific information passes through the unimodal specific encoder. Despite the same information being present in both modalities, they likely represent different aspects, leading to the learning of synergistic effects in the end.

In conclusion, based on our findings, early fusion appears to be more suitable when dealing with completely synergistic modalities. However, the late fusion approaches may still learn substantial cross-modal interactions if redundancies exist.

\begin{table}[t!]

\centering
\begin{adjustbox}{width=\linewidth}
\begin{tabular}{lcccccccccccc} 
\toprule
   & {$\text{Uniqueness }$}  & Synergy & Redundancy  \\
 \midrule

early &  0.2 $\pm$ 0.1 &  98.0 $\pm$ 0.5 &   38.6 $\pm$ 0.5    \\
intermediate & 0.6 $\pm$ 0.3   & 94.8 $\pm$ 0.5 & 47.2 $\pm$ 2.1   \\
late &0.5 $\pm$ 0.2     & 0.5 $\pm$ 0.0 &  1.1 $\pm$  0.4 \\
\bottomrule
\end{tabular}
\end{adjustbox}
\caption[Effects of fusion methods of cross-modal interactions on synthetic data]{Effects of fusion methods using FCNN of cross-modal interactions on HD-XOR datasets with two modalities. }
\label{tab:HD-XOR fusion stratagies}
\end{table}

\paragraph{F1 Score Results.} 

Table \ref{tab:F1macroHDXOR} presents the results for the HD-XOR dataset with two modalities, showing that the model successfully learned the dataset’s underlying logic, as indicated by high F1 macro scores across all settings—uniqueness, synergy, and redundancy—except for the random setting. When we refer to F1 scores, we are specifically referring to F1 macro scores. Notably, learning synergy was the most challenging, with an F1 score of 92.4\% $\pm$ 0.3, while uniqueness was the easiest, achieving a score of 99.5\% $\pm$ 0.1.

\begin{table}[ht!]
\centering

\begin{adjustbox}{width=\linewidth}
\begin{tabular}{lccccccc} 
\toprule
&{{Synergy}} & {{Uniqueness}}& {{Redundancy}}&  Random\\

\midrule
Modality 1 & 50.3 $\pm$  0.0 & 99.5 $\pm$  0.1 &  96.7 $\pm$  0.2 & 49.6 $\pm$ 0.1 \\

 Modality 2 & 51.8 $\pm$  0.0 & 49.3 $\pm$  0.1 & 91.4 $\pm$  0.5 & 51.0  $\pm$ 0.2 \\ 
\midrule
Early  & 92.4 $\pm$ 0.3   & 99.5  $\pm$  0.0  & 97.4  $\pm$ 0.1 & 50.0 $\pm$   1.1 \\  

Intermediate &    80.0  $\pm$ 1.5  & 96.0    $\pm$  0.6   &  92.4  $\pm$ 0.4 & 46.5  $\pm$ 1.9    \\
Late   & 49.7  $\pm$ 1.2  & 60.5  $\pm$ 1.3   &  57.8   $\pm$ 4.5   &  47.3  $\pm$ 1.9  \\

\bottomrule
\end{tabular}

\end{adjustbox}
\caption{F1 macro results for multi- and unimodal FCNN on HD-XOR with two modalities.}

\label{tab:F1macroHDXOR}

\end{table} 

To assess whether each dataset contains the desired amount of synergy, we examine the F1 scores for models trained on individual modalities. A high F1 score in this context suggests that the necessary information to solve the task is present in that modality. 
In the uniqueness setting, the modality containing the relevant information consistently achieved a high F1 score, confirming successful dataset creation. In the redundancy setting, performance was similar for models trained on each modality individually, with improved F1 scores when both modalities were used, indicating effective dataset creation.
For the synergy setting, unimodal inputs performed no better than random, but performance improved significantly when both modalities were used, demonstrating that the synergy dataset was also effectively created.

\subsection{Healthcare Data}
\label{app:mimic3-additional-results}
\paragraph{F1 Score Results.} 

The F1 macro scores for both real datasets for all models used and unimodal models are shown in Table \ref{tab:F1scores single cell} and \ref{tab:F1 MIMICIII}.

\begin{table}[ht!]
\centering

\begin{tabular}{lccccccccc} 
\toprule

 & Single-Cell \\
 \midrule
Modality 1 (Protein) & 59.9 $\pm$ 2.5  \\

Modality 2 (RNA) & 64.3 $\pm$  1.6\\
\midrule
Early &  71.3 $\pm$  8.4 \\
Intermediate &  65.0 $\pm$  0.9  \\
Late &  45.4 $\pm$ 1.1  \\
\bottomrule
\end{tabular}
\caption{F1 score for Mulitmodal Single-Cell dataset.}
\label{tab:F1scores single cell}
\end{table}

\begin{table}[ht!]
\centering

\begin{tabular}{lcc} 
\toprule
&\multicolumn{2}{c}{MIMIC III} \\
\cmidrule(lr){2-3}
 & Task 1  & Mortality \\
 \midrule
Modality 1  & 80.7  $\pm$  0.0   & 17.8 $\pm$  0.5 \\

Modality 2 & 52.9  $\pm$ 1.3  & 16.5 $\pm$ 0.4  \\
\midrule
baseline &   56.8  $\pm$  0.0   & 15.5 $\pm$  0.1   \\
MVAE & 58.0  $\pm$ 0.0  & 15.7  $\pm$ 0.1 \\
\bottomrule
\end{tabular}
\caption{F1 score for MIMIC III dataset.}
\label{tab:F1 MIMICIII}
\end{table}
\end{document}